# NLDF: Neural Light Dynamic Fields for Efficient 3D Talking Head Generation

**Niu Guanchen**[1] , **Li Teng**[2] , **Third Author**[2,3] and **Fourth Author**[4]

[1]ahu unversity
[2]Second Affiliation
[3]Third Affiliation
[4]Fourth Affiliation
{first, second}@example.com, third@other.example.com, fourth@example.com

## Abstract

Talking head generation based on the neural radiation fields (NeRF) model has shown promising visual effects. However, the slow rendering speed of NeRF seriously limits its application, due to the burdensome calculation process over hundreds of sampled points to synthesize one pixel. In this work, a novel Neural Light Dynamic Fields (NLDF) model is proposed aiming to achieve generating high quality 3D talking face with significant speedup. The NLDF represents light fields based on light segments, and a deep network is used to learn the entire light beam's information at once. In learning the knowledge distillation is applied and the NeRF based synthesized result is used to guide the correct coloration of light segments in NLDF. Furthermore, a novel active pool training strategy is proposed to focus on high-frequency movements, particularly on the speaker mouth and eyebrows. The propose method effectively represents the facial light dynamics in 3D talking video generation, and it achieves approximately 30 times faster speed compared to state-of-the-art NeRF based method, with comparable generation visual quality. Code is available at https://github.com/XXX/XXX.

## 1 Introduction

Learning from short video sequences, audio-driven talking head synthesis is a promising research topic with wide-ranging applications such as film production, video conferencing, human-computer interaction, and digital human creation [Wang *et al.*, 2021; Zhang *et al.*, 2020; Zhou *et al.*, 2020].

In recent years, the neural radiation fields (NeRF) model has shown the ability to produce excellent lighting and color effects in 3D volume rendering [Yao *et al.*, 2022; Guo *et al.*, 2021; Shen *et al.*, 2022], which drives the exploration of NeRF based 3D talking head generation. Approaches like ADNeRF [Guo *et al.*, 2021] and DFRF [Shen *et al.*, 2022] can generate high-quality talking videos with simple input training videos. Despite of the promising visual quality, the slow rendering speed limits their applications seriously. To render a single pixel, the NeRF framework requires synthesizing radiance over hundreds of sampled points through alpha compositing, which involves hundreds of network forward passes. For example, to generate a 512×512 resolution 30-second video takes about 7 hours using a V100 GPU with ADNeRF. This prolonged processing time makes it impractical to be widely adopted in real-world applications.

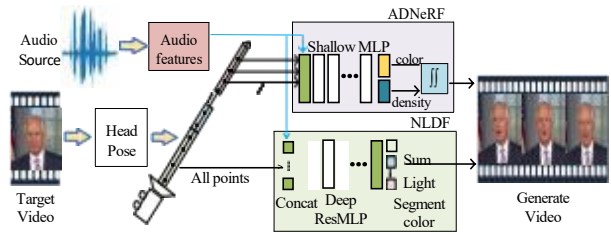

Figure 1: Illustration of our proposed NLDF framework for talking video generation. We utilize head poses extracted from the video to generate light beams. Unlike NeRF based methods, we feed the complete light beam features into a deep residual MLP network, where the output represents color values for light segments along the beams. The final pixel values in the video are obtained by accumulating the color values across all light beam segments.

To ensure high-quality output and fast generation speed simultaneously poses a significant challenge to NeRF based methods, and has attracted several research interests. One intuitive way to improve the rendering speed is reducing the size of the Multi-Layer Perceptrons (MLP) in NeRF [Mildenhall *et al.*, 2021]. Some works focused on sampling points selection and reduction on NeRF beams. For example, [Lin *et al.*, 2022] proposed a depth-guided sampling strategy to enhance the rendering efficiency. However, this method requires extracting depth from perspective images surrounding the object, which does not align with the single-view direction prevalent in the talking video. RAD-NeRF [Tang *et al.*, 2022] introduced a new grid framework necessitate difficult-to-obtain additional features to correct the network. Though effective, these methods did not consider the specific talking video generation task, and compromised the generated video's quality with only limited speedup.

This paper explores in light fields representation perspective and proposes a novel Neural Light Dynamic Fields (NLDF) to speedup the 3D talking head generation task while maintaining the high visual quality. The proposed NLDF renders pixel values in a single pass based on light segments, bypassing the extensive computation of voxel rendering based on many sampling points used in NeRF. Differently from static scene synthesis, in talking head generation task the target videos are usually from specific viewpoints rather than various new perspectives. Therefore the burdensome sequentially fusing every sampling points on the light beam can be optimized by directly inputting the light beam vector representation into the prediction network, along with audio features. To maintain the rendering visual quality, a deep network with extensive residual MLP blocks is adopted in pixel prediction.

It's worth noting that, in 3D talking head generation natural human facial dynamics especially the mouth movement is crucial for effects. To accurately model the light dynamics, We represent a light beam by several light segments. This allows for a comprehensive representation of the speaker's movements by outputting color for each light segment. A fusion of density and color outputs from a teacher NeRF model is used to guide the NLDF learning. Furthermore, a novel active pool training strategy is proposed for effective facial dynamics learning.

Fig. 1 illustrates the proposed framework comparing with ADNeRF. The NLDF achieves rendering speed that is ~30 times faster than ADNeRF and ~70 times faster than DFRF for single-image rendering. It also surpasses previous methods in terms of generation quality in 3D talking video synthesis. In summary, this paper makes the following contributions:

- We propose a novel NLDF model that can generate high-quality 3D talking video driven by audio features efficiently. NLDF renders based on light segments, and it exhibits a significant improvement in rendering speed compared to NeRF based methods. The proposed method also obtains superior generation visual quality.

- To address the challenge of learning dynamics of speaker mouth movements in talking video, we propose a knowledge distillation method to guide the learning of NLDF model with the density and color outputs of the neural radiation fields. The color values of beam segments are estimated by a voxel rendering calculation formula, to handle complex facial movements.

- We propose a novel active pool training strategy to focus on high-frequency movements, particularly on the mouth and eyebrows. This effectively handles the background variation in talking video and accelerates the model's convergence.

## 2 Related work

Existing talking head video generation methods can be divided to two main categories according the resulted visual effects they aim to achieve: 2D-based and 3D-based.

### 2.1 2D-based methods

Early conventional approaches often utilized GAN networks [Chen *et al.*, 2019;Blanz, ;Thies *et al.*, 2016], as well as VAE [Zhou *et al.*, 2019;Eskimez *et al.*, 2021] to generate the 2D face effect with less consideration of 3D head motion. These methods often utilized 2D facial landmarks as the middle representation in generating talking videos. [Zhang *et al.*, 2022] proposed a method by constructing a phoneme-posture dictionary and training a GAN model to synthesize videos from interpolated phoneme postures. [Hegde *et al.*, 2021] introduced a novel speech enhancement paradigm, leveraging the latest advancements in speech-driven lip synthesis. [Zhou *et al.*, 2021] used non-aligned original facial images, employing only one photo as an identity reference, and designed an implicit low-dimensional posture encoding to modularize audio-visual representations.

Recent work [Kim *et al.*, 2022] achieved speaker head generation using latent diffusion models [Ho *et al.*, 2020; Saharia *et al.*, 2022], demonstrating a coexistence of high-quality generation and high generalization. However, without 3D head modeling or radiation fields rendering, the motion naturalness and lighting effects of the result video are still insufficient for user experience.

### 2.2 3D-based methods

To pursuit highly realistic and flexible effects, 3D-based talking head generation attracted much research attention. [Zhang *et al.*, 2021a] utilized a facial 3D morphable model 3DMM [Ma *et al.*, 2023;Zhang *et al.*, 2023] to represent the speaker head and designed a novel optical flow-guided video generator to synthesize videos. [Lahiri *et al.*, 2021] decomposed facial representation into a standardized space, decoupling 3D geometry, head pose, and texture. [Li *et al.*, 2021] proposed a 3D facial model-guided attention network, taking animation parameters as input and using attention masks to manipulate facial expression variations for input individuals. These methods synthesize the talking head via an explicit 3D representation and simulate the three-dimensional features of the face in video generation.

Motivated by the excellent rendering effect of NeRF, some recent methods tried to generate the 3D talking head based on NeRF without any intermediate face representation. ADNeRF [Guo *et al.*, 2021] is the pioneer in introducing NeRF into the domain of talking video generation, achieving promising quality in the resulted content. DFRF [Shen *et al.*, 2022] achieved sample-efficient video generation by learning prior facial knowledge. LipNeRF [Chatziagapi *et al.*, 2023] synchronized lip movements with speech more effectively by combining NeRF with GAN techniques. Addressing the limitations of previous NeRF based methods in generalizing to out-of-domain audio, GENEFACE [Ye *et al.*, 2023] proposed a deformation based approach that can generate natural outcomes corresponding to various out-of-domain audio inputs.

Despite the remarkable achievements, the slow rendering speed still challenges NeRF based methods significantly. In the domain of static scene reconstruction, Instant-NGP [Müller *et al.*, 2022]üemploys hash encoding and utilizes a smaller MLP network to accomplish the reconstruction task, resulting in a significant improvement in rendering speed. However, in the case of speaker NeRF dealing with dynamic scenes and relying on audio as a driving factor, a smaller MLP becomes challenging to ensure effective learning of au-dio features. RAD-NeRF [Tang *et al.*, 2022] accelerates the rendering process of NeRF by reducing the sampling points and utilizing grid encoding. However, this approach requires additional explicit handling of eye features, and does not directly enable learning of blinking movements through the network. This may introduce inconvenience in practical usage.

Our work falls into the 3D-based category and proposes a novel NLDF model to inherit the high-quality rendering effects of NeRF with greatly accelerated generation speed. Unlike previous methods focusing on optimizing NeRF sampling points [Lin *et al.*, 2022; Tang *et al.*, 2022], we represent light beams differently and design specifically for talking facial dynamics, providing a new perspective to 3D talking video generation.

## 3 Method

The proposed NLDF talking video generation framework, depicted in Figure 2, leverages pose parameters estimated from each frame in the video to derive light beam information. This information includes light beam origin $(x_o, y_o, z_o)$ and direction $(x_d, y_d, z_d)$. The audio information corresponding to the image is processed through a feature extraction module to obtain audio features, denoted as **a**. We feed the light beam features processed by the beam encoder module along with audio features **a** into the NLDF model, which comprises an deep residual MLP network.

To accurately represent dynamic scenes, especially in generating precise active mouth pixels, we do not task the network with directly predicting color values. Instead, the network's output comprises various color components for segments of the light beam. This approach aims to enhance the representative capability for dynamic scenes. Considering that the total color value of a light beam can be viewed as the sum of all photons along the light beam, we roughly segment the spatial positions of photons along the light beam direction into $M$ categories. In other words, the light beam is divided into $M$ segments. The network output comprises color values $\{c_{\varphi i}\}$ for all segments of the light beam. The pixel value $c_\varphi$ corresponding to the light beam is derived by accumulating $\{c_{\varphi i}\}$ along the light beam. The final result picture can be generated by computing the pixel values corresponding to all light beams in the image.

The learning of color values $\{c_{\varphi i}\}$ for light beam segments relies on guidance from predicted color values $\{\tilde{c_{ii}}\}$ of light beam segments by the teacher network. In essence, we input the sampling points $(x, y, z)$ and direction $d$ from the corresponding light beam into the trained teacher radiation fields network. Through the teacher model, we predict the **color** and **density** of the sampling point. Based on the classification of light beam segments, voxel rendering is performed separately on $M$ segments of the light beam to calculate the light beam segment color values $\{\tilde{c_{ii}}\}$ by the teacher model. For further details, refer to Subsection 3.3.

### 3.1 NLDF network

**Network design.** The NLDF model consists of a Beam encoder and an 88-layer ResMLP network. The purpose of the beam encoder is to extract and represent light beam information. Deep networks often require a large amount of data for training. In the talking video generation, more than 25 frames per second are needed to maintain continuity in videos. Learning from several minutes video, we can access to thousands of images, providing ample data for the training of NLDF.

By inputting data $(x_o, y_o, z_o, x_d, y_d, z_d)$ representing the light beam and the processed audio feature **a**, we obtain the color values corresponding to various light segments through the following formula:

$$F_\phi : (x_o, y_o, z_o, x_d, y_d, z_d, a) \rightarrow \{c_{\varphi i}\}_{i=1,...M} \quad (1)$$

where $\phi$ represents the trained NLDF model, **a** represents the inputted audio feature, and $c_{\varphi i}$ represents the color value of the i-th segment of the light beam. $M$ represents the number of segments for a single light beam, serving as a hyperparameter, set to 4 in this study. The network produces color values for $M$ light beam segments, which are subsequently accumulated through a rendering process to compute pixel values. In Figure 2, we utilize an 88-layer ResMLP network, as our color prediction network. Since we transform the input of the network from point features to light beam features, a deeper network is needed comparing to NeRF based methods to learn the intricate light beam information.

### 3.2 Feature extraction

**Light beam features.** In NLDF, accurately describing light beam features is crucial for reconstructing frontal-view portraits of individuals in videos. In R2L [Wang *et al.*, 2022], the entire light beam is fed into prediction network to achieve a significant speedup. In this work, although we have employed the approach of using light beam features as the network input, our method differs from R2L. To address the modeling of dynamic scenes, we expect the network not only to learn the color values of the input light beams but also, through knowledge distillation, guide the NLDF in capturing the color distribution along the input light beams.

For rays passing through the spatial domain of the light fields which originate from the camera origin, we represent a light beam using spatial coordinates of $K$ sampling points and concatenate them into an input vector. This $3K$-dimension vector, describing the light beam, is then fed into the NLDF network. Specifically, as shown in Figure 2, the light beams can be represented by the camera origin $(x_o, y_o, z_o)$ and the light beam direction $(x_d, y_d, z_d)$. These raw light beams can be encoded with $K$ sampling points, resulting in features with $3K$ dimension. More sampling coordinate points lead to more precise light beam description. In this work, we choose 16 coordinate points to describe a light beam, aiming to enable the network to capture light beam features more accurately.

**Audio features.** In talking video generation, audio is the driving factor for facial. Therefore the audio feature is part of input for NLDF, deviating from the traditional time-based dynamic field. As depicted in Figure 2, additional audio feature channels are introduced in the establishment of NLDF. To this aim, audio features are extracted by utilizing a pre-trained DeepSpeech [Hannun *et al.*, 2014] model based on RNN. To ensure the smoothness of audio features in the video, for a specified frame, we input its audio features along with those from adjacent frames into the DeepSpeech network. The extracted multi-frame audio features are then processed using an attention-based feature fusion model, resulting in refined audio features **a**. These processed audio features **a** serve as conditional inputs driving the motion of the speaker in resulted videos. Ultimately, the NLDF network also learns the mapping between audio features and mouth movements.

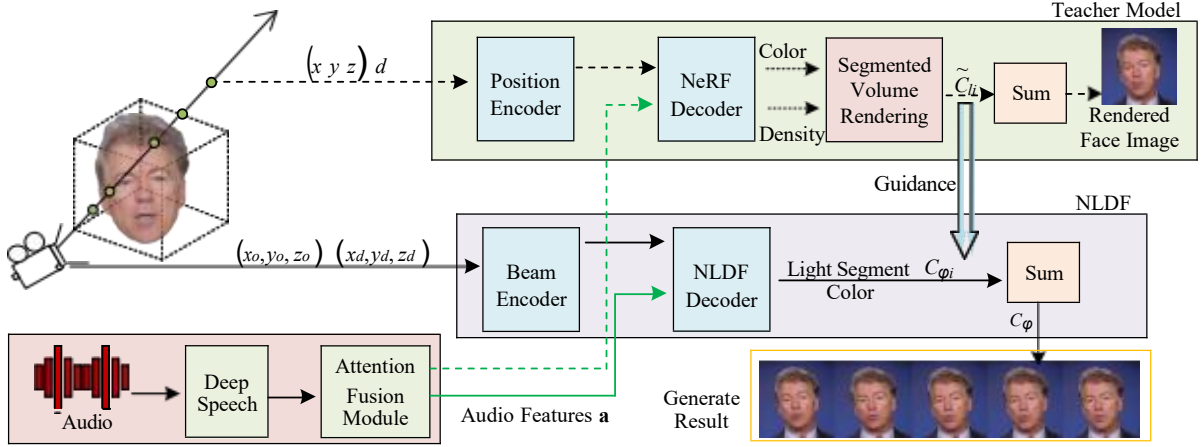

Figure 2: Our framework is primarily divided into three parts: firstly, the audio section, where audio feature extraction utilizes Baidu's deepspeech and an attention fusion module. Secondly, the NeRF based teacher network is employed to guide the proposed NLDF to predict accurate colors for light beam segments. The third part is our NLDF network, where facial poses are encoded into light beam features by the beam encoder. These light beam features, along with audio features, are collectively input into a deep network, producing color values for each light beam segment. By cumulatively summing the color values of each light segment on the light beam, the final generated result is obtained.

### 3.3 NeRF-based teacher network

We introduced a teacher speaker NeRF network to guide the learning of NLDF. As well known, the generated results of the radiation field can to some extent present the distribution of light particles and color values on the light beam. Their information can be used to guide our NLDF in learning light segment representations. The light beam is divided into segments, representing different regions such as the front face, the front half of the head, the back half of the head, and the back of the light beam. This allows NLDF to use multiple light segments to generate color values, instead of a single beam, promoting the learning of high-frequency color changes in the mouth and eyes.

The teacher speaker radiance fields network generates color values using voxel rendering. Following is the network calculation formula for the speaker NeRF, where audio a serves as one driving feature for generating talking videos:

$$F'_\theta : (a, d, p) \rightarrow (c, \sigma) \quad (2)$$

Here, $\theta$ represents the network parameters of the speaker radiance fields, $a$ denotes the audio features, $d$ is the direction of the ray, and $p$ represents the spatial coordinates of the sampling points in the scene. The network outputs color values $c$ and $\sigma$ density along the ray. We followed the network training methodology of ADNeRF [Guo *et al.*, 2021], employing the MSE loss function to train the teacher speaker radiation fields network until convergence and subsequently freezing its network parameters.

### 3.4 Voxel rendering for NLDF

The voxel rendering for the teacher speaker radiance fields network is formulated as follows:

$$C(r, a, \theta, \Pi) = \int_{t_n}^{t_f} T(t)\sigma(r(t))c(r(t), d)\, dt \quad (3)$$

$C$ denotes the accumulation of colors along the light beam, where $r(t) = o + t \times d$, with $o$ as the camera origin and $d$ as the light beam direction. The near boundary is represented as $t_n$, while the far boundary is $t_f$. $\Pi$ is the estimated rigid pose parameters of the face, represented by the rotation matrix $R \in \mathbb{R}^{3\times3}$ and translation vector $t \in \mathbb{R}^{3\times1}$, i.e., $\Pi = \{R, t\}$.

$$T(t) = \exp\left(-\int_{t_n}^{t} \sigma(r(s))\, ds\right) \quad (4)$$

where $T(t)$ is the accumulated transmittance along the ray from $t_n$ to $t$.

We partition the sampling points along the ray direction in the speaker's radiation fields network into $M$ (4 in this study) sets, each containing $N/M$ sampling points (the total number of sampling points in the radiation fields $N$ is four times $N/M$). We apply the voxel rendering formula (Eq.5 below) to calculate the color value $\tilde{c}_{\tilde{i}i}$ of this light segment for each set of sampling points, which is used to guide the color values $c_{\varphi i}$ of the light segments output by the student speaker NLDF network.

$$\tilde{c}_{\tilde{i}i} = \sum_{k=1+(i-1)\times N/M}^{i\times N/M} \exp\left(-\sum_{j=1}^{k-1} \sigma_j \delta_j\right)(1 - \exp(-\sigma_j \delta_j)) c_k$$

$$(5)$$

Here, i represents the subset index of sampling points on the light beam, where i ranges within {1, 2, 3, 4}. $\tilde{c}_{li}$ represents the color value of the i-th light segment calculated by the teacher speaker radiance fields through voxel rendering. σ is the radiance fields sampling point density, and δ is the sampling depth interval. $c_k$ represents the color predicted by the network for the k-th sampling point.

We use the 4 sets of light segment color values obtained from the teacher network to guide the learning of the light segment representation in the student network. The same light beams are inputted into a teacher network, generating corresponding color values c and density values σ. Using voxel rendering, the teacher network predicts color values $\tilde{c}_{li}$ for each segment of the light beams.

In other words, a loss function $loss_{rs}$ is established using $\tilde{c}_{li}$ and $c_{\varphi i}$ of the light segments output by the radiance fields $\tilde{c}_{li}$ to guide the output of the speaker NLDF:

$$loss_{rs} = \sum_{r \in N_l} \sum_{i=1}^{M} \|c_{\varphi i} - \tilde{c}_{li}\|_2^2 \quad (6)$$

where $N_l$ represents the total number of sampled light beams in a training batch. $c_{\varphi i}$ represents the color value of the i-th light segment output by the NLDF network. By aggregating the colors of each light segment on a light beam, the corresponding pixel value for the light beam can be obtained, by the below equation:

$$c_\varphi = \sum_{i=1}^{M} c_{\varphi i} \quad (7)$$

where $c_\varphi$ represents the pixel color values predicted by the NLDF network.

By aggregating the $c_{\varphi i}$ values for each light beam, $loss_r$ is calculated by comparing it against the ground truth pixel values $c_g$.

We calculate the Mean Squared Error (MSE) loss between $c_\varphi$ and the real pixel value $c_g$ from the image. This serves as the $loss_r$ function in learning NLDF:

$$loss_r = \sum_{r \in N_l} \|c_\varphi - c_g\|_2^2 \quad (8)$$

The final loss is defined as follows:

$$loss = loss_r + \lambda loss_{rs} \quad (9)$$

λ is a hyperparameter used during training, and in our experiments, it is set to 0.2.

### 3.5 Active light beam training strategy

As well-known, a major challenge in 3D talking video generation is the active movement of the mouth and eyebrows, followed by facial movements, while the background remains static. To enable the network to effectively capture this characteristic, we propose an active light beam training strategy designed to focus more on active and rapidly changing light beams, specifically the parts with intense movement in the images. For each frame in the image, we aim to have the network focus more on the speaker's facial region.

For each training iteration, 4096 randomly sampled pixels are chosen from an image in the training set, where 90% of these pixels correspond to facial regions. From these pixels, light beam information is derived based on pose parameters, yielding data on the starting point $(x_o, y_o, z_o)$ and direction $(x_d, y_d, z_d)$ of each light beam. This light beam information is fed into a beam encoding module to extract light beam features. Simultaneously, audio features **a** representing the frame corresponding to the image are processed and integrated with the light beam features, forming inputs for a NLDF network.

During each iteration of model training, we sort the loss of each light beam sample in the light beam pool in ascending order and select the top j light beams (a predefined percentage constant) to add or replace in the active beam pool. The beams in the active beam pool are added as extra batches to the model training in each iteration. Note that these additional training batches only optimize our light dynamic field network, and do not affect the audio feature extraction network. The proposed active beam strategy effectively accelerates the convergence of the network, as demonstrated in the following Experiments.

## 4 Experiments

### 4.1 Experimental setup

**Dataset.** The previous ADNeRF [Guo *et al.*, 2021] used high-resolution videos from natural scenes, while the DFRF [Shen *et al.*, 2022] collected speaker video clips from YouTube. We followed the previous works' route and selected high-resolution YouTube videos sourced from the HDTF dataset [Zhang *et al.*, 2021b], which comprises 362 videos from recent YouTube releases, totaling 15.8 hours in length. We selected 10 videos from this dataset, cropped the facial regions, and then resampled the cropped videos to a fixed frame rate of 25 FPS. For each video, the first 80 % is used as the training set and the remaining portion is used as the test set, ensuring no overlap between the training and test sets.

**Head pose.** Since our NLDF aims to generate the reconstruction of the entire facial region, rather than just mouth movements, similar to the approach taken by ADNeRF, we sampled the Face2Face [Thies *et al.*, 2016] method to estimate head poses.

**Evaluation metrics.** Our proposed method will be evaluated based on quantitative metrics combined with visual results. Common metrics for evaluating the effectiveness of generated radiance fields include Peak Signal-to-Noise Ratio (PSNR↑), Structural Similarity Index (SSIM↑) [Hore and Ziou, 2010], and Learned Perceptual Image Patch Similarity (LPIPS↓) [Wang *et al.*, 2004]. For a generated video, we observed that the PSNR↑ tends to assign higher scores to blurry images with lower generation accuracy [Zhang *et al.*, 2018]. In contrast, LPIPS↓ aligns better with human perception. SyncNet scores (offset↓/confidence↑) [Chung and Zisserman, 2017] are used to assess the synchronization quality between audio and visual cues in the generated videos, which is crucial in the speaker generation domain.

| Knowledge distillation | PSNR↑ | SSIM↑ | LPIPS↓ |
|---|---|---|---|
| Without | 32.38 | 0.9539 | **0.0837** |
| With | **32.59** | **0.9553** | 0.0852 |

Table 1: Knowledge distillation ablation experiment.

| Active strategy | | PSNR↑ | SSIM↑ | LPIPS↓ |
|---|---|---|---|---|
| With | 25k | 30.27 | 0.9671 | 0.1315 |
|  | 50k | **30.37** | 0.9647 | 0.1175 |
|  | 100k | 30.02 | 0.9621 | **0.1065** |
| Without | 25k | 30.13 | **0.9679** | 0.1390 |
|  | 50k | 30.28 | 0.9654 | 0.1241 |
|  | 100k | 30.22 | 0.9640 | 0.1145 |

Table 2: Active strategy ablation experiment.

**Training details.** All our experiments were conducted on a Linux system using two V100 GPUs. The project code is based on PyTorch [Paszke *et al.*, 2019]. We employed the Adam optimizer [Kingma and Ba, 2014] to optimize both the light fields model and the audio feature extraction network.

### 4.2 Ablation experiment

**Knowledge distillation.** In dynamic speaker fields, directly inferring precise pixel values from light beams using the network proves challenging, often resulting in blurriness around areas like the mouth and eyes. Therefore, we employ a teacher radiation fields network to guide the student network in approximating the color distribution along the light beams. This approach significantly improves the generation quality of active motion areas, as depicted in Figure 3. In images generated without this method, pixel distortions occur around the mouths of the characters. While utilizing this method, the generated images exhibit better mouth shapes. Metrics from ablation experiments are presented in Table 1. While the improvements in metrics are moderate, they contribute to rectifying frames with blurry artifacts.

**Active light beams training strategy.** During model training, the network often converges quickly in non-active areas, e.g. the background. The proposed active training strategy aims to improve the computational efficiency, and expedite model convergence. Table 2 demonstrates the effectiveness of this strategy by comparing two classes of models, with and without this strategy, after 25k, 50k, and 100k training iterations. Minimal differences in PSNR↑ and SSIM↑ scores are observed across the same number of iterations, indicating the successful establishment of fundamental character portraits by the models. To accurately depict the convergence status of the models, we use LPIPS↓ metrics, which exhibit higher scores for models employing this strategy at the same iteration counts. Furthermore, LPIPS↓ scores progressively improve with increased iteration counts, suggesting that this strategy enhances the convergence quality of the models.

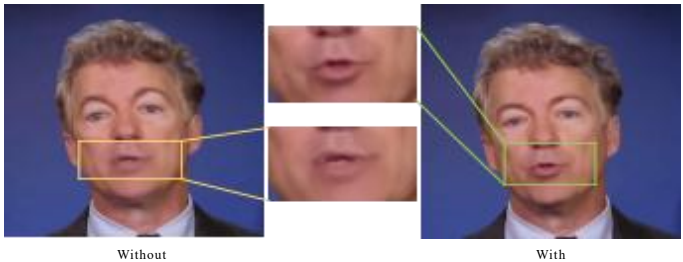

Figure 3: Comparative comparison of knowledge distillation ablation experiments.

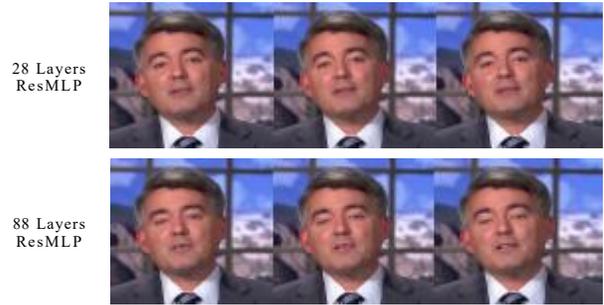

Figure 4: Network depth of the capture of eyeblink movements.

| Network-depth | PSNR↑ | SSIM↑ | LPIPS↓ |
|---|---|---|---|
| 28 layer | **30.23** | **0.9618** | 0.1076 |
| 50 layer | 30.06 | 0.9592 | 0.0996 |
| 88 layer | 29.95 | 0.9557 | **0.0926** |

Table 3: Network-depth contrast experiment.

**Network depth.** In this subsection, we explore the expressive capabilities of different network depths in NLDF model. We evaluate the generation performance metrics for converged models with network depths of 28, 50, and 88 layers, as presented in Table 3. It's observable that increasing the number of layers doesn't lead to significant differences in PSNR↑ and SSIM↑ scores but results in noteworthy improvements in LPIPS↓ scores. Importantly, as depicted in Figure 4, we observed that the 28-layer network model struggles to capture blinking movements, often depicting wide-open eyes in the generated videos. With the increase in network depth, the capability to capture blinking movements improves, and at 88 layers, it can effectively generate blinking motions. This illustrates that shallow networks in our NLDF method are insufficient to depict complete facial movements. Increasing network depth enhances the network's ability to learn facial movements, underscoring the importance of deep networks within our NLDF method.

### 4.3 Comparison of methods

**Comparison with NeRF-like methods.** In this section, we compared our method with NeRF-like methods using two different speaker target datasets. The evaluation results presented in Table 4 were obtained after approximately 600k training iterations, ensuring the convergence of the models. We generated videos with a frame rate of 25 frames per second, resulting in 28-second speaker videos for both targets (A and B). As presented in Table 4, our NLDF exhibits slightly higher scores in PSNR↑ and LPIPS↓ compared to ADNeRF and approaches proximity to DFRF. However, DFRF holds a slight edge over us in SyncNet scores. As depicted in Figure 5, our NLDF method closely aligns with both approaches, showcasing excellent generation quality. On the same V100 GPU, for generating a single frame image of video at a resolution of 768×768, our method requires approximately 1 second. In contrast, ADNeRF takes around 28 seconds, and DFRF requires approximately 70 seconds. Our proposed method exhibits significantly faster generation speed than both of them.

|  | Test set A | | | | | Test set B | | | |
| --- | --- | --- | --- | --- | --- | --- | --- | --- | --- |
| Method | PSNR↑ | SSIM↑ | LPIPS↓ | SyncNet↓↑ | Flops Speedup | PSNR↑ | SSIM↑ | LPIPS↓ | SyncNet↓↑ |
| GT |  |  |  | -6/2.471 |  |  |  |  | -5/8.967 |
| ADNeRF | 27.91 | **0.960** | 0.1255 | -15/0.466 | 1× | 24.75 | **0.9557** | 0.1462 | -10/0.776 |
| DFRF | 31.94 | 0.954 | 0.0897 | **-6/1.516** | ~0.4× | 29.39 | 0.9522 | **0.0858** | **-5/5.544** |
| Ours | **32.59** | 0.955 | **0.0852** | -6/1.416 | ~30× | **29.95** | 0.9557 | 0.0926 | -5/4.267 |

Table 4: Comparison of the NeRF class methods."GT" stands for the real videos corresponding to the generated videos.It's preferable for the SyncNet scores offset↓ to be close to the GT scores.

|  | Test set A | | | | Test set B | | | |
| --- | --- | --- | --- | --- | --- | --- | --- | --- |
| Method | PSNR↑ | SSIM↑ | LPIPS↓ | SyncNet↓↑ | PSNR↑ | SSIM↑ | LPIPS↓ | SyncNet↓↑ |
| GT |  |  |  | -6/2.471 |  |  |  | -5/8.967 |
| Makeittalk | 21.24 | **0.986** | 0.1649 | -15/0.088 | 16.13 | **0.965** | 0.2272 | -11/0.311 |
| Wav2lip | 29.86 | 0.960 | 0.1190 | -3/**2.934** | 28.19 | **0.965** | 0.1286 | -3/**8.842** |
| Ours | **32.59** | 0.955 | **0.0852** | -6/1.416 | **29.95** | 0.956 | **0.0926** | -5/4.267 |

Table 5: Compared with the 2D method."GT" stands for the real videos corresponding to the generated videos.It's preferable for the SyncNet scores offset↓ to be close to the GT scores.

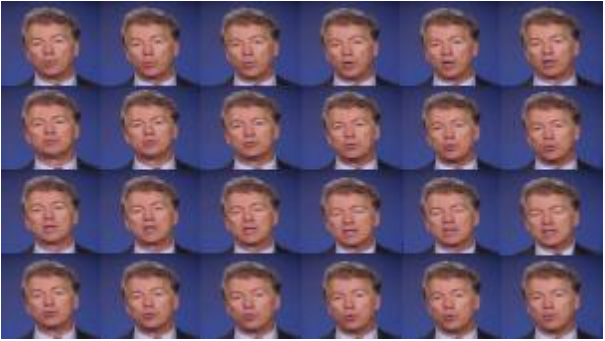

Figure 5: From top to bottom, there are consecutive images of real videos (GT), followed by images generated by DFRF, ADNeRF, and our method, respectively, for six frames.

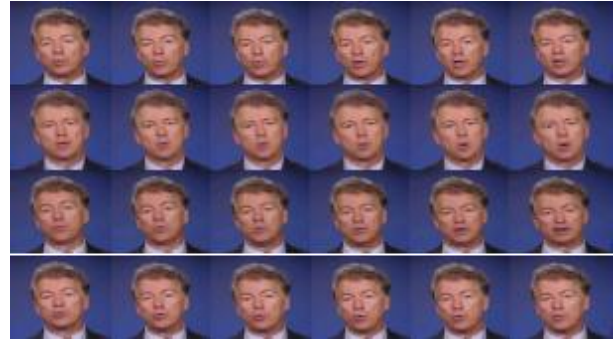

Figure 6: From top to bottom, there are consecutive images of real videos (GT), followed by images generated by Makeittalk, Wav2lip, and our method, respectively, for six frames.

**Comparison with 2D based methods.** We further compared our method with previous popular 2D methods on the same datasets, and the results are shown in Table 5. Unlike our approach, Wav2lip [Prajwal et al., 2020] only generates mouth pixels and exhibits blurriness in the generated images, resulting in subpar generation quality. Similarly, Makeittalk [Zhou et al., 2020] generates faces that inaccurately represent facial features of individuals, as depicted in Figure 6. Wav2lip achieves higher sync scores due to its usage of Sync-Net as a discriminator, but it exhibits differences from natural real video mouth shapes. Figure 6 illustrates that the mouth shapes generated by these 2D methods resemble a generic shape, whereas our method generates mouth shapes based on a specific individual used during model training, making them closer to the mouth shapes observed in real videos.

## 5  Conclusion

In this paper, we propose a novel NLDF method for achieving high-fidelity generation of talking head animation. Our approach directly synthesizes human head movements from audio signals and supports free manipulation of head poses. Guided by knowledge distillation, our trained NLDF model demonstrates an exceptionally rapid rendering speed without compromising the precision of image generation. The rendering speed improvement is approximately 30 times faster than speaker radiation fields methods, marking a substantial advancement in the practical application of virtual digital humans. Additionally, we implemented an active beam training strategy to expedite model convergence and enhance computational efficiency.